\documentclass[journal,10pt,twocolumn,twoside]{IEEEtran}
\IEEEoverridecommandlockouts                              % This command is only
\usepackage{mathtools}
\usepackage{graphicx}
\usepackage{pgfplots}
	\newlength\figureheight
	\newlength\figurewidth
	\pgfplotsset{compat=newest}
	\usepgfplotslibrary{external}
	\usepgfplotslibrary{groupplots}
    \usetikzlibrary{external} % Moved to main
\usepackage{subcaption}
\usepackage[utf8]{inputenc}
\usepackage[T1]{fontenc}
\usepackage{xcolor}
\usepackage{amsfonts}
\usepackage{url}
\title{\LARGE \bf
Visualising Deep Network's Time-Series Representations
}

\author{Błażej Leporowski and Alexandros Iosifidis\\
Department of Engineering, Aarhus University, Denmark\\
\{bl, ai\}@eng.au.dk}

\begin{document}

\maketitle
\thispagestyle{empty}
\pagestyle{empty}

%%%%%%%%%%%%%%%%%%%%%%%%%%%%%%%%%%%%%%%%%%%%%%%%%%%%%%%%%%%%%%%%%%%%%%%%%%%%%%%%
\begin{abstract}
Despite the popularisation of machine learning models, more often than not, they still operate as black boxes with no insight into what is happening inside the model.
There exist a few methods that allow to visualise and explain why a model has made a certain prediction.
Those methods, however, allow visualisation of the link between the input and output of the model without presenting how the model learns to represent the data used to train the model as whole.
In this paper, a method that addresses that issue is proposed, with a focus on visualising multi-dimensional time-series data.
Experiments on a high-frequency stock market dataset show that the method provides fast and discernible visualisations.
Large datasets can be visualised quickly and on one plot, which makes it easy for a user to compare the learned representations of the data.
The developed method successfully combines known techniques to provide an insight into the inner workings of time-series classification models.

\end{abstract}
%%%%%%%%%%%%%%%%%%%%%%%%%%%%%%%%%%%%%%%%%%%%%%%%%%%%%%%%%%%%%%%%%%%%%%%%%%%%%%%%

\section{Introduction}
On average, 21 million trades per day take place on 9 stock exchange markets managed by NASDAQ Inc. alone \cite{NASDAQ}. 
What this number illustrates is that every day an enormous amount of financial data is generated around the world. 
This data represents a wide range of information: stocks, commodities, complex industry markers, mortgage trackers, etc. 
The shared characteristic of all those different types of data is that they are all represented as time-series. 
Various combinations of this immense amount of data are constantly analysed by both human experts and machines. 
Many machine learning models have been applied to such datasets \cite{tabl,Kercheval2015,Passalis2019,Makinen2018,Niu2020,Nousi2018,Qin2017,Passalis2019a}. 
Most often only the predictions that these models produce are visualised, without providing any insight into what is happening inside the model.

Visualisation serves an important role in the process of digesting and gaining knowledge from this unceasing stream of information \cite{VanWijk2006}. 
It provides an efficient tool for tasks such as time-series comparison \cite{Fulcher,Fulcher2014}, outlier detection \cite{Hyndman2016}, process understanding \cite{Kang2017} or providing insights into the information that the time-series hold \cite{Nguyen2017}. 
The methods proposed in \cite{Fulcher,Fulcher2014,Hyndman2016,Kang2017} lose their relevancy when combined with pre-trained (off-the-shelf) Deep Learning models. 
This is because these methods aim to discover patterns and trends in the raw time-series data, but they do not consider the data transformations and the internal data representations learned by high-performing Deep Learning models. 
Employing a time-series visualisation approach that is disconnected from the model used for predictions can lead to misleading conclusions. 
It is thus important to create a data visualisation model that can be attached to any Deep Learning model trained for time-series classification and allow for easy visualisation of its internal representations for new time-series data. 

Moreover, an efficient approach for visualising internal data representations of Deep Learning models can provide useful insights during model selection. 
Often, a wide collection of Deep Learning models is used to tackle a problem and the best performing one is chosen. 
While there is a range of metrics that are used to evaluate the quality of a model, without including visualisation in this process the information that a human expert receives in order to make a decision is incomplete. 
Visualising how well each model has learned to represent classes forming the classification problem, combined with the additional quality metrics such visualisation can provide, can aid the process of decision making.

In this paper, we propose a data visualisation method that enables visualisation of time-series representations that Deep Learning models learn. 
Visualising those representations as $2$- or $3$-dimensional data points enables models of different types and structures to be easily compared. 
To account for the possible impact of various processing steps on the final visualisations, the proposed method targets all steps in the process, i.e. from time-series classification to the final visualisation.  
The resulting visualisation provides an insight into how the Deep Learning model transforms the input time-series data in order to classify it and allows a human expert to gain useful insight.
The contributions of our work are:
\begin{itemize}
    \item We propose a time-series data visualisation method that exploits information encoded in the parameters of a trained Deep Learning model.  
    \item We combine our visualisation method with Deep Learning models used for time-series data from a high-frequency stock market. 
    We provide both quantitative and qualitative comparisons of the different Deep Learning models based on their classification performance and the quality of the intrinsic time-series data representations, respectively. 
    \item We observe that the quality of visualisations of the intrinsic time-series data representations can give an indication of the classification performance of the different models.
\end{itemize}

The remainder of the paper is structured as follows. 
Section \ref{S:Method} describes the proposed method and Section \ref{S:Data} describes the data used in our experiments for evaluating the method when combined with state-of-the-art Deep Learning models for time-series classification. 
Section \ref{S:Experiments} describes our experiments and Section \ref{S:Results} provides the obtained results. Finally, conclusions are drawn in Section \ref{S:Conclusions}.

%%%%%%%%%%%%%%%%%%%%%%%%%%%%%%%%%%%%%%%%%%%%%%%%%%%%%%%%%%%%%%%%%%%%%%%%%%%%%%%%
\section{Method}\label{S:Method}
The goal of the proposed method is to provide a visualisation of the time-series data representations at a hidden layer of a Deep Learning model used for time-series classification.
In order to do this, we use a dimensionality reduction method that receives as input the time-series data representations at a hidden layer of the Deep Learning model and produces a 2-dimensional representation that can be visualized by the user.

Let us express the data transformations performed by a Deep Learning model trained for time-series data classification as 
\begin{equation}
    \mathbf{X}_i \xrightarrow{F(\mathbf{X_i})} \mathbf{z}_i \xrightarrow{C(\mathbf{z}_i)} \mathbf{s}_i, \label{Eq:GenericClassifier}
\end{equation}
where $\mathbf{X}_i = [\mathbf{x}_{i1}, \dots , \mathbf{x}_{i T_i}] \in \mathbb{R}^{D \times T_i}$ is an input time-series formed by $T_i$ vectors $\mathbf{x}_{it} \in \mathbb{R}^D$, each representing the time-series state at time instance $t$. 
$F(\mathbf{X_i})$ is a parametric function that maps the input time series $\mathbf{X_i}$ to a $d$-dimensional vector $\mathbf{z}_i$ and $C(\mathbf{z}_i)$ is the last layer of the Deep Learning model, commonly formed by a linear transformation followed by a softmax activation function mapping $\mathbf{z}_i$ to a probability-like class vector $\mathbf{s}_i \in \mathbb{R}^C$, where $C$ is the number of classes in the time-series classification problem at hand. 
After obtaining $\mathbf{s}_i$, the input time-series $\mathbf{X}_i$ is assigned to the class corresponding to the maximal output. 
State-of-the-art Deep Learning models for time-series classification jointly optimize the parameters of the functions $F(\cdot)$ and $C(\cdot)$ to maximize the probability of correct class assignment on the time-series data of the training set formed by multiple time-series $\{\mathbf{X}_i, l_i\}, \:i=1,\dots,N$, where $l_i \in \{1,\dots,C\}$ is the class label corresponding to $\mathbf{X}_i$. 

Considering the data transformation in Eq. (\ref{Eq:GenericClassifier}), the role of function $F(\cdot)$ is to jointly perform a (hierarchical) nonlinear transformation of the input vectors $\mathbf{x}_{it} \in \mathbb{R}^D, \:i=1,\dots,N, \:t=1,\dots,T_i$ and aggregate the information of each time-series $i$ to form the time-series representation $\mathbf{z}_i \in \mathbb{R}^d$. 
The role of $C(\cdot)$ is to determine a linear classifier in the feature space $\mathbb{R}^d$ determined by $\mathbf{z}_i, \:i=1,\dots,N$. 
That is, in order to visualize the classification results it is sufficient to use the time-series data representations in $\mathbb{R}^d$.
Since in most of the existing Deep Learning models for time-series classification $d > 3$, the dimensionality of the time-series data representation needs to be reduced while preserving as much information and structure of the time-series data representations $\mathbf{z}_i, \:i=1,\dots,N$. 
That is, we would like to learn a mapping $Q(\mathbf{z}_i)$ such that:
\begin{equation}
    \mathbf{X}_i \xrightarrow{F(\mathbf{X_i})} \mathbf{z}_i \xrightarrow{Q(\mathbf{z}_i)} \mathbf{y}_i, \label{Eq:GenericMapper}
\end{equation}
where $\mathbf{y}_i \in \mathbb{R}^2$ (or $\mathbb{R}^3$). The mapping $Q(\cdot)$ can be defined by applying a parametric dimensionality reduction model to allow for easy out-of-sample operation. 
To consider both global and local information of the time-series representations in $\mathbb{R}^d$ for determining this mapping, we define $Q(\cdot)$ as a neural network trained by the parametric t-SNE method \cite{par_tsne}. 

Here we should note that when optimizing for the parameters of the mapping $Q(\cdot)$ in Eq. (\ref{Eq:GenericMapper}), one can freeze the parameters of $F(\cdot)$ and, thus, define $2$- or $3$-dimensional representations of $\mathbf{z}_i, \:i=1,\dots,N$. 
Alternatively, one can finetune the parameters of $F(\cdot)$ while optimizing the parameters of $Q(\cdot)$. 
The advantage of the first case is that the final low-dimensional representations $\mathbf{y}_i$ of the input time-series $\mathbf{X}_i$ correspond to the representations used by the classification layer $C(\cdot)$. 
Thus, visualization of the low-dimensional vectors $\mathbf{y}_i$ can give to human experts insights on the quality of the time-series data representations used for classification. 
The advantage of the latter case is that it can be used to provide more descriptive time-series data representations compared to the first case. 
By initializing the parameters of $F(\cdot)$ to those used by the Deep Learning model in Eq. (\ref{Eq:GenericClassifier}), the low-dimensional time-series data representations can capture information which is important for classification. 
However, by jointly updating the parameters of the functions $F(\cdot)$ and $Q(\cdot)$, the final low-dimensional time-series data representations will not correspond to the original representations $\mathbf{z}_i, \:i=1,\dots,N$ used by the classification layer $C(\cdot)$.

%%%%%%%%%%% Describe classifiers
\subsection{Deep Learning models for time-series classification}
As mentioned above, the proposed time-series data visualization method can be used to visualise the hidden layer representations obtained by any Deep Learning model of the form shown in Eq. (\ref{Eq:GenericClassifier}). 
In this paper, we consider four Deep Learning models that have been successfully applied in financial time-series classification problems.
Each classifier is trained by minimizing the categorical cross entropy loss defined as
\begin{equation}
L(\mathcal{W}) = - \sum_{i=1}^N \sum_{c=1}^{C} t_{ic} \cdot log (s_{ic}),
\end{equation}
where $\mathcal{W}$ denotes the parameters of the Deep Learning model (i.e. the parameters of both $F(\cdot)$ and $C(\cdot)$ functions), $\mathbf{t}_i \in \mathbb{R}^C$ is the (one-hot) target vector defined based on $l_i$ and $\mathbf{s}_i \in \mathbb{R}^C$ is the classifier's output vector for the time-series $\mathbf{X}_i$. $t_{ic}$ and $s_{ic}$ correspond to the $c$-th element of vectors $\mathbf{t}_i$ and $\mathbf{s}_i$, respectively, indicating the prediction of the model for class $c$ given as input the time-series $\mathbf{X}_i$. 
To minimize this loss, a gradient descent-based optimization process is used. In its basic form it can be described as:
\begin{equation}
\mathcal{W} \leftarrow \mathcal{W} - \eta \cdot \frac{\partial L(\mathcal{W})}{\partial \mathcal{W}},
\end{equation}
where $\eta$ is the learning rate. Multiple variants of gradient descent-based optimization exist. We used the Adaptive Moment Estimation (Adam) optimizer \cite{kingma2014adam}, which is based on adaptive estimates of lower-order moments.

\subsubsection{Multilayer Perceptron (MLP)}
MLP \cite{MURTAGH1991183} is a neural network consisting entirely of fully-connected layers (also known as dense layers). Each neuron in a fully-connected layer is connected to all neurons in the previous and next layers and it performs an affine transformation of its inputs. That is, the $k$-th neuron on layer $m$ receives an input vector $\mathbf{x}^{(m-1)} \in \mathbb{R}^{D_{(m-1)}}$, where $D_{(m-1)}$ is equal to the number of neurons of layer $m-1$, and produces an output: 
\begin{equation}
    \mathbf{x}^{(m)}_{k} = \phi\left( \mathbf{W}^{(m)}_{k} \mathbf{x}^{(m-1)} + \mathbf{b}^{(m)}_k \right),
\end{equation}
where $\mathbf{W}^{(m)} \in \mathbb{R}^{D_{(m)} \times D_{(m-1)}}$ and $\mathbf{b}^{(m)}_k$ are the weight matrix and bias vector of layer $m$, respectively. $\phi(\cdot)$ is the activation function nonlinearly transforming the output of each neuron.

\subsubsection{Convolutional Neural Network (CNN)}
CNN \cite{LeCun1999} is formed by a series of convolutional and pooling layers followed by a set of fully-connected layers. 
Each convolutional layer $m$ is equipped with a set of filters $\mathbf{W}^{D_{(m)} \times S \times  D_{(m-1)}}$, where the first dimension denotes to the number of filters in layer $m$, the second dimension denotes the filter size, and the third dimension denotes the number of input channels. 
The output of a convolutional layer can be pooled using a pooling layer aggregating the information of its input to reduce the size of the time-series representation. 
After the last convolutional/pooling layer fully-connected layers are used to classify the input time-series.

\subsubsection{Long- Short-Term Memory (LSTM)}
LSTM \cite{lstm_article} network is a neural network formed by at least one LSTM unit, which is used to jointly transform and aggregate information of an input time-series. 
The LSTM unit is composed of a cell, input gate, output gate and a forget gate. 
The gates regulate the flow of information in and out of the cell. 
For an LSTM unit placed at the $m$-th layer of the LSTM network, the following equations define its response:
\begin{eqnarray}
    \mathbf{f}^{(m)}_t &=& \phi_g \left(\mathbf{W}^{(m)}_f \mathbf{x}^{(m-1)}_{it} + \mathbf{R}^{(m)}_f \mathbf{h}^{(m)}_{t-1} + \mathbf{b}^{(m)}_f \right),  \label{equation:forget_gate}    \\
    \mbox{\boldmath{$\iota$}}^{(m)}_t &=& \phi_g \left(\mathbf{W}^{(m)}_{\iota} \mathbf{x}^{(m-1)}_{it} + \mathbf{R}^{(m)}_{\iota} \mathbf{h}^{(m)}_{t-1} + \mathbf{b}^{(m)}_{\iota} \right),  \label{equation:input_gate}    \\
    \mathbf{\tilde{c}}^{(m)}_t &=& \phi_h \left(\mathbf{W}^{(m)}_c \mathbf{x}^{(m-1)}_{it} + \mathbf{R}^{(m)}_c \mathbf{h}^{(m)}_{t-1} + \mathbf{b}^{(m)}_c \right), \label{equation:cell_input}  \\
    \mathbf{c}^{(m)}_t &=& \mathbf{f}^{(m)}_t  \mathbf{c}^{(m)}_{t-1} + \mbox{\boldmath{$\iota$}}^{(m)}_t \mathbf{\tilde{c}}^{(m)}_t,   \label{equation:cell_state}    \\
    \mathbf{o}^{(m)}_t &=& \phi_g \left(\mathbf{W}^{(m)}_o \mathbf{x}^{(m-1)}_{it} + \mathbf{R}^{(m)}_o \mathbf{h}^{(m)}_{t-1} + \mathbf{b}^{(m)}_o \right),  \label{equation:output_gate}    \\
    \mathbf{h}^{(m)}_t &=& \mathbf{o}^{(m)}_t \phi_h (\mathbf{c}^{(m)}_t),   \label{equation:hidden_state}
\end{eqnarray}
where $\phi_g(\cdot)$ and $\phi_h(\cdot)$ are the sigmoid and tanh activation functions, respectively. 
Matrices $\mathbf{W}^{(m)}_{\times}$ and $\mathbf{R}^{(m)}_{\times}$ correspond to the weights of the input and recurrent connections for the gate type $\times$, i.e. forget ($f$), input ($\iota$) and output ($o$) gate, respectively.

\subsubsection{Temporal Attention-Augmented Bilinear Network (TABL)}
TABL \cite{tabl} network uses bilinear layers to learn two dependencies for the two modes of a multivariate time-series coupled with an attention mechanism that allows the model to focus on the important time instances of the input time-series. 
% A bilinear layer placed at the $m$-th layer of the network transforms its input $\mathbf{X}^{(m-1)}_i \in \mathbb{R}^{D_{(m-1)} \times T_{(m-1)}}$ to an output $\mathbf{X}^{(m)}_i \in \mathbb{R}^{D_{(m)} \times T_{(m)}}$. This transformation is described by:
% \begin{equation}
%     \mathbf{X}^{(m)}_i = \phi \left(\mathbf{W}^{(m)}_1 \mathbf{X}^{(m-1)}_i \mathbf{W}^{(m)}_2 + \mathbf{B}^{(m)} \right),
%     \label{equation:bilinear_layer}
% \end{equation}
% where $\mathbf{W}^{(m)}_1 \in \mathbb{R}^{D_{(m)} \times D_{(m-1)}}$ and $\mathbf{W}^{(m)}_2 \in \mathbb{R}^{T_{(m-1)} \times T_{(m)}}$ are weights and $\mathbf{B}^{(m)} \in \mathbb{R}^{D_{(m)} \times T_{(m)}}$ are biases.
When applied to time-series, this mapping learns two separate transformations of the input time-series, one for transforming the input dimensions and the second aggregating information in the time domain.
To automatically determine which time instances should be emphasized when aggregating information in the time domain an attention mechanism is incorporated in the bilinear mapping leading to the Temporal Attention-Augmented Layer (TABL). 
A forward pass through a TABL layer placed at the $m$-th layer of the network is described by the following equations:
\begin{eqnarray}
    \bar{\mathbf{X}}^{(m)}_i &=& \mathbf{W}^{(m)}_1 \mathbf{X}^{(m-1)}_i, \label{equation:TABL_1} \\
    \mathbf{E}^{(m)} &=& \bar{\mathbf{X}}^{(m)}_i \mathbf{W}^{(m)},    \label{equation:TABL_2} \\
    a^{(m)}_{ij} &=& \frac{e^{e^{(m)}_{ij}}}{\sum_{k=1}^{T_{(m)}} e^{e^{(m)}_{ik}}},    \label{equation:TABL_3} \\
    \tilde{\mathbf{X}}^{(m)}_i &=& \lambda \left(\bar{\mathbf{X}}^{(m)}_i \odot \boldsymbol{A} \right) + (1 - \lambda) \bar{\mathbf{X}}^{(m)}_i,  \label{equation:TABL_4} \\
    \mathbf{X}^{(m)}_i &=& \phi \left(\tilde{\mathbf{X}}^{(m)}_i \mathbf{W}^{(m)}_2 + \boldsymbol{B}^{(m)} \right),    \label{equation:TABL_5}
\end{eqnarray}
where the attention is expressed through the values $a^{(m)}_{ij}$. 
The introduction of attention mechanism in the TABL layer encourages the competition between neurons representing different temporal steps of the same dimension.

%%%%%%%%%%% Describe dimensionality reduction methods
\subsection{t-distributed Stochastic Neighbor Embedding}
The t-distributed Stochastic Neighbor Embedding (t-SNE) \cite{tsne} is a nonlinear dimensionality reduction method. 
Given high-dimensional data set $\mathbf{z}_i \in \mathbb{R}^{d}$, $i=1,\dots,N$, t-SNE aims to create its low-dimensional representation $\mathbf{y}_i \in \mathbb{R}^{2}$ (or $\mathbb{R}^3$), $i=1,\dots,N$. 
To achieve this, t-SNE creates pairwise probability matrices for all data points in both the high-dimensional space $\mathbb{R}^{d}$ and the low-dimensional space $\mathbb{R}^{2}$ (or $\mathbb{R}^3$).
Each probability $p_{i|j}$ represents the probability of point $\mathbf{z}_i$ choosing point $\mathbf{z}_j$ as its neighbour if neighbours were picked in proportion to their probability density under a Gaussian distribution centered at $\mathbf{z}_i$.
% This probability is calculated by:
% \begin{equation}
%     p_{j|i}=\frac{\frac{e^{-||\mathbf{z}_i-\mathbf{z}_j||^2}}{2\sigma_i^2}}{\sum_{k \neq i}\frac{e^{-||\mathbf{z}_i-\mathbf{z}_k||^2}}{2\sigma_i^2}}, \quad for \quad i \neq j
%     \label{equation:t-sne_probability}
% \end{equation}
% and represents the similarity of datapoints $\mathbf{z}_i$ and $\mathbf{z}_j$. 
% The variance $\sigma_i$ centered over each high-dimensional dimensional data point $\mathbf{z}_i$ influences the width of the Gaussian distribution. 
% The higher the value, the flatter the distribution is.
The perplexity parameter in t-SNE expresses a measure of the number of neighbours each point will take into account when calculating the probability matrices.

% Two matrices, $\boldsymbol{P}\in \mathbb{R}^{N \times N}$ and $\boldsymbol{U}\in \mathbb{R}^{N \times N}$ that hold the probabilities of the data pairs being neighbours in the high-dimensional and low-dimensional spaces are constructed. The positions of the samples $\mathbf{y}_i$ are then decided by minimizing the Kullback-Leibler divergence of the distribution $\boldsymbol{P}$ over the distribution $\boldsymbol{U}$. 
% The KL divergence for $\boldsymbol{P}$ and $\boldsymbol{U}$ becomes:
% \begin{equation}
%     KL(\boldsymbol{P}||\boldsymbol{U})=\sum_{i \neq j}p_{ij}log\frac{p_{ij}}{u_{ij}}. \label{Eq:KL_par_tSNE}
% \end{equation}
% By minimizing the KL divergence using gradient descent a map reflecting the high-dimensional inputs in low-dimensional space is obtained.

The parametric variant of the t-SNE \cite{par_tsne} aims to provide a solution for out of sample, non-linear dimensionality reduction. 
Parametric t-SNE employs a fully-connected neural network to learn the parametric mapping $Q(\mathbf{z}_i) = \mathbf{y}_i$ between high dimensional data $\mathbf{z}_i, \:i=1,\dots,N$ and their lower-dimensional representations $\mathbf{y}_i, \:i=1,\dots,N$.
% by optimizing for $KL(\boldsymbol{P}||\boldsymbol{U})$ in (\ref{Eq:KL_par_tSNE}).

%%%%%%%%%%%%%%%%%%%%%%%%%%%%%%%%%%%%%%%%%%%%%%%%%%%%%%%%%%%%%%%%%%%%%%%%%%%%%%%%
\section{Limit Order Book data}\label{S:Data}
The FI-2010 dataset \cite{benchmark_dataset} is a public dataset of high-frequency Limit Order Books for mid-price prediction.
A Limit Order Book (LOB) is a record of limit orders.
A limit order can be of two types, i.e., bid or ask.
The bid order aims to buy at a specified price or lower. 
The ask order aims to sell at a specified price or higher. 
The LOB is a record of all those orders.
The stock market then matches the bid orders with the lowest ask order and vice versa.
These matches are made immediately after a new order arrives and the amount of shares in the LOB decreases by the traded amount.
The FI-2010 dataset tracks the mid-price movement specifically. 
The mid-price is the mean between the best bid and ask prices.
As such, no trades take place at mid-price, but it provides a useful metric to track the movement of the market.
The dataset labels the changes in the mid-price into three categories: downward, stable, and upward trends.
More information on limit order books can be found in \cite{lob}.

The FI-2010 dataset contains data from 5 Finnish companies, all operating in different sectors of the industry, listed at NASDAQ Nordic. 
The data was collected in a period from 01.07.2010 to 14.07.2010, producing data spanning 10 working days and around 4.5 million events. 
For each event, the top 10 prices and volumes from both sides of the LOB, i.e. the best 10 bid orders and the best 10 ask orders, are extracted. 
Each event is thus represented by a 40-dimensional vector.
Each 40-dimensional vector is then assigned a label for each of the following five prediction horizons: next 10, 20, 30, 50, and 100 events. 
The data is normalized using z-score normalization. 

In the experimental setup, the first 7 days are used as training data and the last 3 days as test data.
The used prediction horizon is the one corresponding to the next 10 events.
This amounts to 254,651 training  and 139,290 test time-series.

%%%%%%%%%%%%%%%%%%%%%%%%%%%%%%%%%%%%%%%%%%%%%%%%%%%%%%%%%%%%%%%%%%%%%%%%%%%%%%%%
\section{Experiments}\label{S:Experiments}
We conducted experiments by using the four types of Deep Learning models described in Section \ref{S:Method}. 
As a baseline, we use the best performing Deep Learning model combined with Principal Component Analysis to compare the obtained time-series data representations with those obtained by parametric t-SNE. 

The architectures of the four Deep Learning model types were determined by performing a comparative study following the time-series classification experimental protocol provided by the database. 
During this process, we set the value of $d = 60$, and optimized the parameters of each Deep Learning model for $200$ epochs with a batch size of $1024$ and the Adam optimizer with default settings. 
To account for the imbalance between the classes forming the classification problem, we used weighted categorical cross-entropy loss with weights $w_c = \frac{N}{C \cdot N_c}$ where $N$ is the total number of samples, $C$ is the number of classes, and $N_c$ is the number of samples in the c-the class. 
We used an initial learning rate of $0.001$ and applied the learning rate decay schedule with the multiplication factor of $0.5$, a minimal learning rate of $0.0001$, and patience of $4$ epochs.
We used early stop callback with a minimum loss change of $0.0001$ and patience of $6$ epochs, as well as a dropout rate of $10\%$ when it is applicable. This process led to the following neural network architectures:
\begin{itemize}
    \item The MLP receives as input a $40$-dimensional vector and consists of $5$ fully-connected layers with $400$-$800$-$500$-$60$-$3$ neurons, respectively. The first $4$ layers use ReLU activation function and the last one uses softmax activation function. All layers except the last one use Dropout. 
    
    \item The LSTM network receives as input $10$ $40$-dimensional vectors and consists of $2$ LSTM layers, containing $128$ and $60$ units, respectively, and a fully-connected layer with $3$ neurons. The LSTM layers use tanh activation function and the fully-connected layer uses softmax activation function. In the LSTM layers Dropout is used. 
    
    \item The CNN is formed by three $1$-dimensional convolutional layers with $8$-$16$-$32$ filters using ReLU activation function and causal padding with a kernel size of 3, one 1D Max-Pooling layer with size 2, one fully-connected layer with 60 neurons and ReLU activation function receiving as input the flattened output of the pooling layer, and a fully-connected softmax layer with 3 output neurons. 
    
    \item The TABL network was implemented following \cite{tabl} which was thoroughly tested in the problem studied in this paper. It receives as input a matrix of $40 \times 10$ dimensions and it is formed by four layers leading to time-series representations of $60 \times 10$, $120 \times 5$, $60 \times 1$ and $3 \times 1$ representations. All layers use ReLU activation function, except the last one which uses softmax.
\end{itemize}

After determining the best architecture for each Deep Learning model, we dropped the last (classification) layer and combined it with parametric t-SNE. 
To account for the randomness introduced to the end-to-end training process of the combined model, we first optimized the parameters of the parametric t-SNE network (i.e. the mapping $Q(\cdot)$) by fixing the parameters of the  classification network (i.e. the mapping $F(\cdot)$). Subsequently, the parameters of the combined model (i.e. $Q(\cdot)$ and $F(\cdot)$) are jointly finetuned.

The implementation of the parametric t-SNE is based on the publicly available implementation\footnote{\url{https://github.com/jsilter/parametric\_tsne}} of the original paper \cite{par_tsne}. 
The original network architecture has four fully-connected layers formed by $500$-$500$-$2000$-$2$ neurons, and each of these layers is pretrained as an autoencoder. 
In our experiments, we consider several network architectures for the parametric t-SNE by varying the number and size of the fully-connected layers and the perplexity parameter. 
The targets for the t-SNE method take the form of probability matrices constructed for each batch of data. 
These can be precomputed for accelerating the training process significantly. 
We optimize the t-SNE networks for $200$ epochs using a batch size of $2048$ samples. 
Our experiments have shown that smaller batch sizes tend to give worse results. 
Following the implementation of the original t-SNE paper \cite{tsne}, trustworthiness and k-NN classification performance metrics are used to evaluate the quality of the visualisations. 
Trustworthiness expresses how well the local structure of the data has been preserved.

\section{Results}\label{S:Results}
We evaluated different aspects of the proposed method, including the impact of the perplexity parameter and the network architecture on the performance of parametric t-SNE, the effects of end-to-end optimization of the combined model, the relationship between the performance of the classification model and the produced visualisation, and comparisons with end-to-end unsupervised time-series visualisation. We start by reporting the classification performance of the four Deep Learning models in the classification task, followed by the performance of the baseline method.

\subsection{Comparison of Deep Learning models in classification}     \label{section:classifier_comparison}
In our first experiment, we evaluated the performance of the four Deep Learning models in the classification task of the FI-2010 dataset. We used accuracy, precision, recall and f1-score, as shown in \ref{tab:classifiers_comparison}. 
Due to the fact that the FI-2010 dataset describes a class-imbalanced problem, the f1-score was used for comparing the performance of the classification models. 
As can be seen, the TABL network performs the best, and its performance is in line with the performance achieved in the original paper \cite{tabl} (which is $77.63\%$). 
LSTM and CNN networks achieve decent performance and can be expected to produce adequate visualisations. The MLP network fails to take note of the spatial nature of time-series, and as such lags considerably behind networks that are able to do so.
The TABL classifier is used in the following experiments unless stated otherwise, as it performed the best.

\begin{table}[h]
    \caption{Comparison of Deep Learning models in the classification problem of FI-2010 dataset.}
    \label{tab:classifiers_comparison}
    \centering
        \begin{tabular}{l|lll|l}
         Classifier     & Accuracy  & Precision & Recall    & f1-score  \\ \hline
         MLP            & 62.02\%   & 64.05\%   & 62.46\%   & 61.83\%   \\
         CNN            & 70.57\%   & 70.92\%   & 70.61\%   & 70.59\%   \\
         LSTM           & 72.84\%   & 72.82\%   & 72.85\%   & 72.8\%    \\
         TABL           & 76.91\%   & 77.46\%   & 76.7\%    & \bf 76.95\%  % \\
%         Original TABL \cite{tabl}  & 84.7\%    & 76.95\%   & 78.44\%   & 77.63\%   
        \end{tabular}
\end{table}

%%%%%%%%%%%%%%%%%%%%%%%%%%%%%%

\subsection{PCA baseline}  \label{section:PCA_performance}
\begin{figure}[h]
    \begin{center}
        \includegraphics[width=0.85\linewidth]{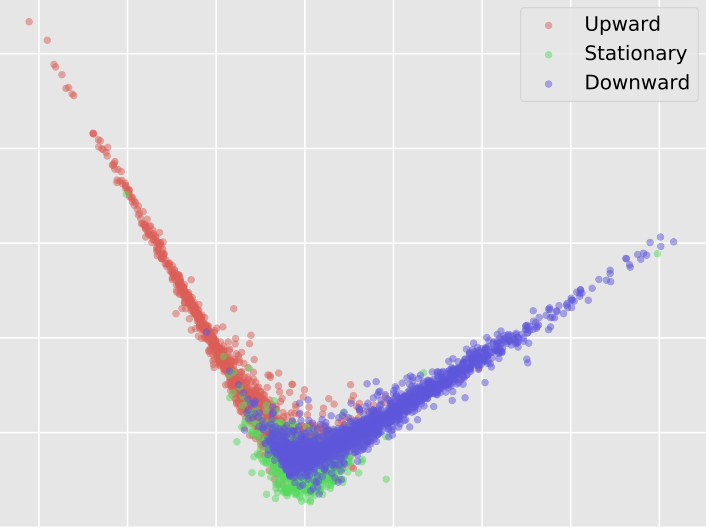}
    \end{center}
    \caption{Visualisation of the combined TABL-PCA model.}
    \label{fig:pca_visualisation}
\end{figure}

In our next experiment, we aim to establish a baseline visualisation by applying Principal Component Analysis to map the time-series representations from $\mathbb{R}^d$ to $\mathbb{R}^2$. 
The resulting visualisation achieves a trustworthiness score of $81.7\%$, which means that the subspace obtained by applying PCA manages to capture the local structure of the dataset well.
Applying a 3-NN classifier on the resulting $2$-dimensional time-series representation leads to an accuracy of $65.56\%$. 
As can be seen, the performance metrics achieved by PCA are relatively high. 
However, the resulting visualization is poor. 
This is because, as can be seen in Figure \ref{fig:pca_visualisation}, the projection spans the two axes of highest variance explained by samples belonging to two classes, while a large percentage of samples from all three classes collapse to the same area. 
Thus, no clusters can be identified, and the visualised samples lie very close to each other. 
This visualisation does not provide much insight about the model's representation of the predictions. 
All visualisations present a subset of $7,500$ samples randomly selected from the test data.
Presenting all $139,290$ points would result in a plot too densely populated to provide any meaningful insight. 

%%%%%%%%%%%%%%%%%%%%%%%%%%%%%

\subsection{Perplexity and parametric t-SNE architecture impact}   \label{section:perplexity_impact}
To evaluate the effect of different combinations of perplexity and parametric t-SNE network architecture on its performance, an extensive search has been conducted. Results for different architectures are shown in Table \ref{tab:perplexity_results}.

\begin{table}[h]
    \caption{Comparison of TABL-t-SNE visualisation architectures.
    Each architecture is presented as an array $[l_1, ..., l_n]$, where $l_i$ is the number of neurons in each layer of the parametric t-SNE network.}
    \label{tab:perplexity_results}
    \centering
        \begin{tabular}{ll|cc}
         Perplexity     & Architecture           & Trustworthines      & 3-NN score \\ \hline
         $5$            & [$500, 500, 2000, 2$]  & $76.17\%$           & $70.17\%$  \\
         $10$           & [$500, 500, 2000, 2$]  & $76.27\%$           & $70.42\%$  \\
         $25$           & [$500, 500, 2000, 2$]  & $76.22\%$           & $70.33\%$  \\
         $50$           & [$500, 500, 2000, 2$]  & $76.05\%$           & $70.14\%$  \\
         $100$          & [$2$]               & $75.22\%$           & $\mathbf{71.00\%}$  \\
         $100$          & [$250, 500, 2$]        & $76.30\%$           & $69.69\%$  \\
         $100$          & [$250, 1000, 2$]       & $76.26\%$           & $69.99\%$  \\
         $100$          & [$500, 2000, 2$]       & $76.52\%$           & $69.62\%$  \\
         $100$          & [$100, 100, 500, 2$]   & $76.20\%$           & $69.47\%$  \\
         $100$          & [$250, 250, 1000, 2$]  & $76.39\%$           & $70.10\%$  \\
         $100$          & [$500, 500, 2000, 2$]  & $\mathbf{79.53\%}$  & $69.89\%$  \\
         $1000$         & [$500, 500, 2000, 2$]  & $76.55\%$           & $70.41\%$  \\
         $2000$         & [$500, 500, 2000, 2$]  & $76.47\%$           & $70.49\%$  \\
         $5000$         & [$500, 500, 2000, 2$]  & $76.39\%$           & $70.48\%$  \\
        \end{tabular}
\end{table}

It is worth noting that the optimization problem used to determine the parameters of the parametric t-SNE network is not convex. 
Thus, it is expected to lead to slightly different results for different weight initialisation. 
To account for this effect, we ran three experiments for each combination and report the average performance scores in Table \ref{tab:perplexity_results}. 
As shown by the experiment performed with PCA, the quality of the visualisation cannot be evaluated by these metrics alone. 
The choice of the best combination is thus a matter of balancing the best performance scores and the best provided visualisation, which is a subjective judgment. 

The impact of perplexity on the performance of the visualization model was shown to be small. 
With the differences in the performance scores being small and after taking into consideration the quality of the obtained visualisations, the perplexity value of $100$ was chosen as the one leading to the best performance. 
The architecture with 3 hidden layers formed by $500$, $500$, $2000$, and $2$ neurons, as proposed in the original paper, has performed the best and was chosen for the following experiments. 
%%%%%%%%%%%%%%%%%%%%%%%%%%%%%%%

\subsection{Impact of end-to-end visualization learning} \label{section:finetuning_impact}
As has been mentioned before, the combined neural network can be trained either by freezing the parameters of the classification model or by performing end-to-end training of the entire model. 
Here we evaluate the impact of end-to-end optimization on the performance of the model. 
To account for the randomness introduced to the end-to-end training process, we first optimize the parameters of the parametric t-SNE network by fixing the parameters of the classification model and, subsequently, the parameters of the combined model are jointly finetuned. 
Results are shown in Table \ref{tab:finetunig_impact}. 
\begin{table}[h]
    \caption{Comparisons in performance based on the low-dimensional time-series representations.}
    \label{tab:finetunig_impact}
    \centering
        \begin{tabular}{l|cc}
         Mode           & Trustworthiness    & 3-NN score \\ \hline
         Finetuned      & 79.53\%           & 69.89\%\\
         Not finetuned  & 69.69\%           & 51.54\%\\
        \end{tabular}
\end{table}

\begin{figure}[h]
    \centering
    \begin{subfigure}{0.5\linewidth}
        \centering
        \includegraphics[width=0.95\linewidth]{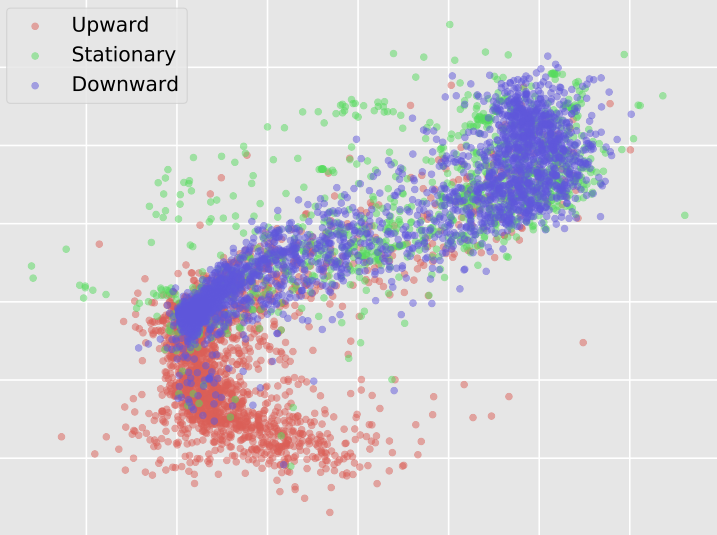}
        \caption{Not finetuned}
        \label{subfig:not_finetuned}
    \end{subfigure}%
        \begin{subfigure}{0.5\linewidth}
        \centering
        \includegraphics[width=0.95\linewidth]{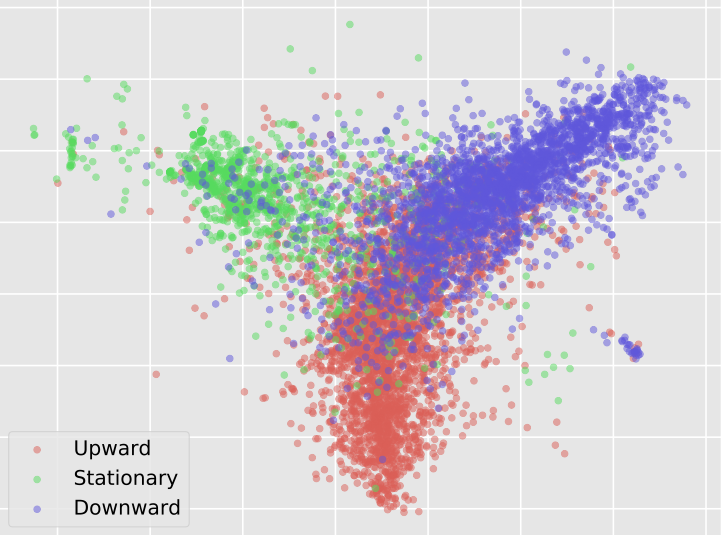}
        \caption{Finetuned}
        \label{subfig:finetuned}
    \end{subfigure}
    \caption{Visualisation of the time-series data obtained by using the original classification model (left) and the classification model fine-tuned jointly with the visualisation model (right).}
    \label{fig:finetuning_visualisation}
\end{figure}

With $9.84\%$ improvement in trustworthiness and $18.35\%$ improvement in $3$-NN score, it is clear that finetuning the model leads to significant performance improvement in the low-dimensional space where visualization is conducted. 
This difference is also visible in the visualisations shown in Figure \ref{fig:finetuning_visualisation}. 
As can be seen in Figure \ref{subfig:finetuned}, the clusters of upward, stationary and downward classes are better separated than in Figure \ref{subfig:not_finetuned}.

Compared to the results achieved using PCA, trustworthiness is lower by $2.17\%$, while the $3$-NN accuracy is higher by $4.33\%$.
In terms of metrics, those two variants achieve comparable results. 
However, a significant difference is visible in the obtained visualisation. 
The visualisation obtained by PCA does not lead to a good clustering of the data and results in a very cluttered plot. 
Visualisations obtained by parametric t-SNE plot the samples in more well-distinguishable clusters and allow for better discrimination between different categories of data. 
The finetuning process is thus confirmed to improve the quality of the visualisation.

%\subsection{Classification performance after finetuning}
Table \ref{tab:degradation} presents the impact of the finetuning process on the classification performance. 
To measure the performance of the fineturned classification model, we remove the visualisation network and add a softmax classification layer which is trained on the three classes of the classificaiton problem. 
This allows us to evaluate how much the finetuning process affected the ability of the classification model to perform the original classification task.
As can be seen, across all metrics, the performance falls by around $1\%$ to $1.5\%$ after finetuning. 
This means that the finetuning process does not significantly degrade discrimination ability of the time-series data representations. 
Therefore, the finetuned visualization model produces visualisations based on representations of high discrimination ability.

\begin{table}[h]
    \caption{Impact of finetuning on the classification performance.}
    \label{tab:degradation}
    \centering
        \begin{tabular}{l|cccc}
         Classifier     & Accuracy  & Precision & Recall    & f1 score \\ \hline
         Finetuned      & 75.35\%   & 76.59\%   & 75.45\%   & 75.42\%\\
         Not finetuned  & 76.91\%   & 77.46\%   & 76.7\%    & 76.95\%\\

        \end{tabular}
\end{table}

%%%%%%%%%%%%%%%%%%%%%%%%%

\subsection{Impact of classifier accuracy on visualization}     \label{section:classifier_impact}
\begin{figure*}[h]
    \centering
    \begin{subfigure}[b]{0.32\linewidth}   
        \centering 
        \includegraphics[width=\textwidth]{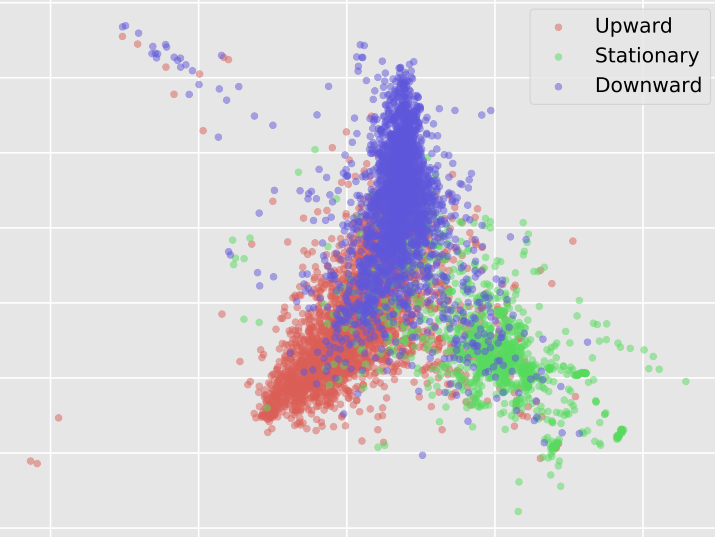}
        \caption[CNN model]%
        {{\small CNN model}}    
        \label{fig:mean and std of net34}
    \end{subfigure}
    \begin{subfigure}[b]{0.32\linewidth}  
        \centering 
        \includegraphics[width=\textwidth]{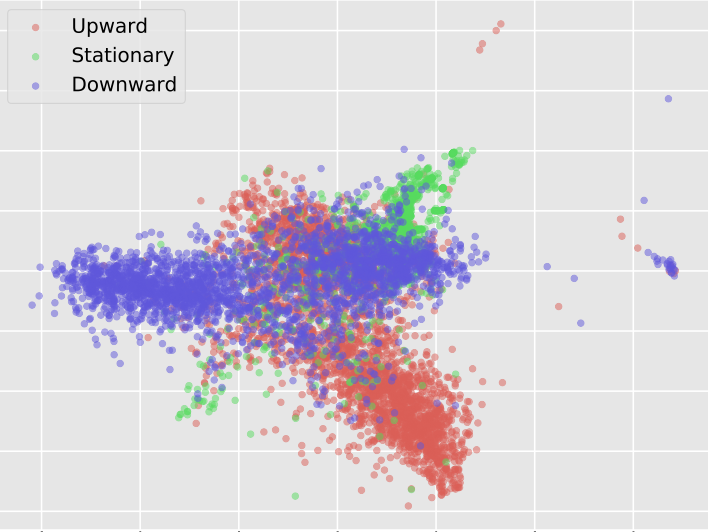}
        \caption[LSTM model]%
        {{\small LSTM model}}    
        \label{fig:mean and std of net24}
    \end{subfigure}
    \begin{subfigure}[b]{0.32\linewidth}
        \centering
        \includegraphics[width=\textwidth]{Test_visualisation_finetuned_TABL.png}
        \caption[TABL model]%
        {{\small TABL model}}    
        \label{fig:mean and std of net14}
    \end{subfigure}
    \caption[Time-series data visualisations based on different classification models.]
    {\small Time-series data visualisations based on different classification models.} 
    \label{fig:classifier_visualisation_comparison}
\end{figure*}

In the previous experiment it has been shown that the finetuned version of the combined model offers a superior performance compared to the model using the original TABL network. 
In this experiment, we compare the performance of the combined networks based on pre-trained MLP, CNN, LSTM and TABL networks and finetuning in terms of trustworthiness, $3$-NN based classification and visualization. 
The results are shown in Table \ref{tab:classifier_impact}. 
Figure \ref{fig:classifier_visualisation_comparison} provides a comparison of the visualisations obtained by the CNN, LSTM and TABL networks. 
We omitted the MLP network, as it was the worse performing model in the classification task. 
\begin{table}[h]
    \caption{Impact of the classification model on visualization performance.}
    \label{tab:classifier_impact}
    \centering
        \begin{tabular}{l|cc}
         Classifier     & Trustworthines       & 3-NN score \\ \hline
         MLP            & $52.97\%$            & $34.77\%$  \\
         CNN            & $60.39\%$            & $69.64\%$  \\
         LSTM           & $75.01\%$            & $64.93\%$  \\
         TABL           & $\mathbf{79.53\%}$   & $\mathbf{69.89\%}$   
        \end{tabular}
\end{table}

The visualizations obtained for the CNN and LSTM networks are worse than the one obtained by using TABL with PCA in terms of metrics. 
A comparison of the corresponding visualisations, however, shows that the visualisations obtained for the CNN and LSTN networks are superior compared to the one obtained by combining TABL with PCA. 
TABL, when combined with parametric t-SNE, outperforms all other models. 

It is worth pointing that even though the CNN-based model falls behind the TABL-based model in terms of trustworthiness, its $3$-NN score is very similar to the one achieved by TABL. 
The visualisation produced by the CNN-based model is also outpefromed only by the one produced by the TABL-based model. 
The low trustworthiness score, however, makes the resulting visualisation not a faithful representation of local structure in the original data. 
The LSTM based model captures the middle ground, preserving well the local structure in the dataset, but not performing very well in terms of cluster discrimination. 
From Tables \ref{tab:classifiers_comparison} and \ref{tab:classifier_impact} it can be seen that the better the classification model is, the better the quality of visualisation is produced. 
However, there is no clear relationship between the performance of the classification model and the quality of the visualisation it produces.

%%%%%%%%%%%%%%%%%%%%%%%%%%

\subsection{Comparisons with end-to-end unsupervised time-series visualization}
Finally, we compare the proposed method in visualizing the internal time-series representations of classification models with  low-dimensional time-series representations obtained by end-to-end training of networks based on parametric t-SNE, in a similar manner as in \cite{Nguyen2017}. 
Table \ref{tab:random_initialization} presents a performance comparison between the finetuned network based on a previously trained classifier and parametric t-SNE networks and a network with the same architecture that has been trained directly by using the parametric t-SNE criterion (\ref{Eq:KL_par_tSNE}). 
In order to fairly compare the two approaches, we ran the end-to-end unsupervised version of the parametric t-SNE network $10$ times and report the mean performance value and the corresponding standard deviation in this Table. 

\begin{table}[h]
    \caption{Visualization performance comparison between supervised initialization and end-to-end unsupervised training.}
    \label{tab:random_initialization}
    \centering
        \begin{tabular}{l|cc}
        Mode                    & Trustworthiness    & 3-NN score \\ \hline
        Full training           & $79.53\%$         & $69.89\%$ \\
        Random initialization   & $55.77\pm0.04\%$ & $35.83\pm0.06\%$  \\
        \end{tabular}
\end{table}

\begin{figure}[h]
    \begin{center}
        \includegraphics[width=0.95\linewidth]{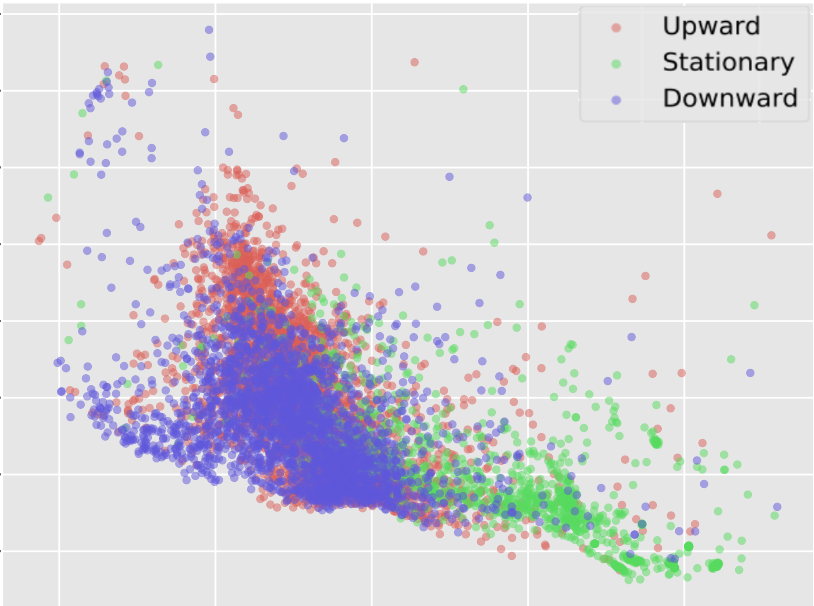}
    \end{center}
    \caption{Visualisation produced by following an unsupervised training process.}
    \label{fig:random_visualisation}
\end{figure}

It is clear that end-to-end training a time-series visualization network in an unsupervised manner is not effective in the financial time-series data used in our experiments. 
This difference in quality is even more visible when considering the visualisation produced by the best performing unsupervised t-SNE network, as seen in Figure \ref{fig:random_visualisation}. 
Comparing the visualization in Figure \ref{fig:random_visualisation} with that in Figure \ref{fig:classifier_visualisation_comparison}c, we can see that time-series representations obtained by the TABL model lead to better discrimination of the samples forming the three classes. This conclusion is in line with the large difference in performance provided in Table \ref{tab:random_initialization} for the two cases. 

%%%%%%%%%%%%%%%%%%%%%%%%%%%%%

%%%%%%%%%%%%%%%%%%%%%%%%%%%%%%%%%%%%%%%%%%%%%%%%%%%%%%%%%%%%%%%%%%%%%%%%%%%%%%%%
\section{Conclusions}\label{S:Conclusions}
In this paper, we proposed a time-series visualization method that is able to provide high-quality visualizations of the internal time-series representations obtained by Deep Learning classification models. 
The visualisations obtained by the proposed method give an intuitive and trustworthy view of how a classification model has learned to represent the input time-series data for making predictions, and how data representations of different classes differ from one another. 
By testing and comparing several classification models of varying complexity it is possible to gain new knowledge about the behaviour of different neural network architectures. 
This knowledge can help to address the deficiencies in the process of selecting an appropriate classification model for the task at hand. 
By visualising and comparing the performance of various classification models in an intuitive way the proposed method can also help to evaluate whether the classification model behaves in an expected and desirable way.

A limitation of the proposed method is linked to the amount of data it is tasked to visualise. That is, visualisation of tens of thousands of time-series data in a low-dimensional feature space can lead to a cluttered and unreadable plot. To avoid such a situation, visualisation of a smaller and representative fragment of the dataset is advised. 
Possible future research directions based on the proposed method include: a) exploitation of the time-series data representations provided by the classification model over over successive training epochs for guiding its training process, b) the creation of multi-scale time-series data visualisations which can be obtained by connecting time-series representations obtained by multiple layers of the classification model, and c) creation of hybrid $3$-dimensional data visualisations corresponding to the $2$-dimensional visualisation of the model's response over the instances of the input time-series (third dimension).
%%%%%%%%%%%%%%%%%%%%%%%%%%%%%%%%%%%%%%%%%%%%%%%%%%%%%%%%%%%%%%%%%%%%%%%%%%%%%%%%
%\section*{Appendix}
%\input{Chapters/Appendix}
%%%%%%%%%%%%%%%%%%%%%%%%%%%%%%%%%%%%%%%%%%%%%%%%%%%%%%%%%%%%%%%%%%%%%%%%%%%%%%%%
\section*{Acknowledgment}
Alexandros Iosifidis acknowledges funding from the project DISPA (grant 9041-00004B) funded by the Independent Research Fund Denmark.
%%%%%%%%%%%%%%%%%%%%%%%%%%%%%%%%%%%%%%%%%%%%%%%%%%%%%%%%%%%%%%%%%%%%%%%%%%%%%%%%
%% Bibliography
% \addtolength{\textheight}{-11.5cm}  % This command serves to balance the column lengths
%                                   % on the last page of the document manually. It shortens
%                                   % the textheight of the last page by a suitable amount.
%                                   % This command does not take effect until the next page
%                                   % so it should come on the page before the last. Make
%                                   % sure that you do not shorten the textheight too much.

\newpage
\bibliographystyle{IEEEtran}
\bibliography{references}
%\printbibliography
%\renewcommand*{\bibfont}{\raggedright}

\end{document}